\pdfoutput=1

\documentclass[11pt]{article}

\usepackage[final]{ACL}

\usepackage{times}
\usepackage{latexsym}

\usepackage[T1]{fontenc}

\usepackage[utf8]{inputenc}

\usepackage{microtype}

\usepackage{inconsolata}
\usepackage{multicol}
\usepackage{tablefootnote}
\usepackage{booktabs}
\usepackage{colortbl}
\usepackage[fleqn]{amsmath}
\usepackage{multirow}
\usepackage[normalem]{ulem}
\useunder{\uline}{\ul}{}
\usepackage{hyperref}
\usepackage{graphicx}
\usepackage{subfigure}
\usepackage{fixltx2e}
\usepackage{xcolor}
\usepackage{algorithm}
\usepackage{algorithmic}
\usepackage{adjustbox}
\usepackage{setspace}
\usepackage{amsfonts}
\usepackage[most]{tcolorbox}
\usepackage{afterpage}

\newcommand{\ie}{\textit{i.e.}}
\newcommand{\eg}{\textit{e.g.}}
\newcommand{\ours}{\texttt{e5-omni}}

\title{\ours{}: Explicit Cross-modal Alignment for Omni-modal Embeddings}

\author{Haonan Chen$^*$$^{1}$, Sicheng Gao\thanks{\ \ Co-first authors} \,$^{2}$, Radu Timofte$^{2}$, Tetsuya Sakai$^{3}$, Zhicheng Dou\thanks{\ \ Corresponding authors} \,$^{1}$ \\
         $^1$Gaoling School of Artificial Intelligence, Renmin University of China \\ 
         $^2$University of Würzburg $^3$Waseda University  \\ 
         \texttt{\{hnchen,dou\}@ruc.edu.cn} \\
        \texttt{\{sicheng.gao,radu.timofte\}@uni-wuerzburg.de} \\
         \texttt{tetsuya@waseda.jp} \\
}

\newtcolorbox[list inside=prompt]{prompt}[1][]{
    colbacktitle=black!60,
    coltitle=white,
    fontupper=\footnotesize,
    boxsep=5pt,
    left=0pt,
    right=-1pt,
    top=0pt,
    bottom=0pt,
    boxrule=1pt,
    width=\textwidth,
    #1,
}
\begin{document}
\maketitle

\begin{abstract}
Modern information systems often involve different types of items, \eg, a text query, an image, a video clip, or an audio segment.
This motivates omni-modal embedding models that map heterogeneous modalities into a shared space for direct comparison.
However, most recent omni-modal embeddings still rely heavily on implicit alignment inherited from pretrained vision-language model (VLM) backbones.
In practice, this causes three common issues:
(i) similarity logits have modality-dependent sharpness, so scores are not on a consistent scale;
(ii) in-batch negatives become less effective over time because mixed-modality batches create an imbalanced hardness distribution; as a result, many negatives quickly become trivial and contribute little gradient; and
(iii) embeddings across modalities show mismatched first- and second-order statistics, which makes rankings less stable.
To tackle these problems, we propose \ours{}, a lightweight explicit alignment recipe that adapts off-the-shelf VLMs into robust omni-modal embedding models.
\ours{} combines three simple components:
(1) modality-aware temperature calibration to align similarity scales,
(2) a controllable negative curriculum with debiasing to focus on confusing negatives while reducing the impact of false negatives, and
(3) batch whitening with covariance regularization to better match cross-modal geometry in the shared embedding space.
Experiments on MMEB-V2 and AudioCaps show consistent gains over strong bi-modal and omni-modal baselines, and the same recipe also transfers well to other VLM backbones.
We release our model checkpoint at \url{https://huggingface.co/Haon-Chen/e5-omni-7B}.

\end{abstract}

\section{Introduction}

Modern information retrieval increasingly needs to handle multimodal queries and results in web search, multimedia question answering, and long-document understanding~\citep{MMEB,MMEB-v2,DSE}.
Recent vision--language models (VLMs) have shown strong multimodal capabilities, and bi-modal embedding models fine-tuned from VLMs perform well on text--image retrieval~\citep{mme5,GME}.
However, many real-world retrieval scenarios involve additional modalities beyond text and images, such as audio and video.
Motivated by this, we study an omni-modal setting where heterogeneous items (text, images, audio, and video) are embedded into a shared space for consistent similarity comparison.

\begin{figure}[!t]
	\centering
	\includegraphics[width=1.0\linewidth]{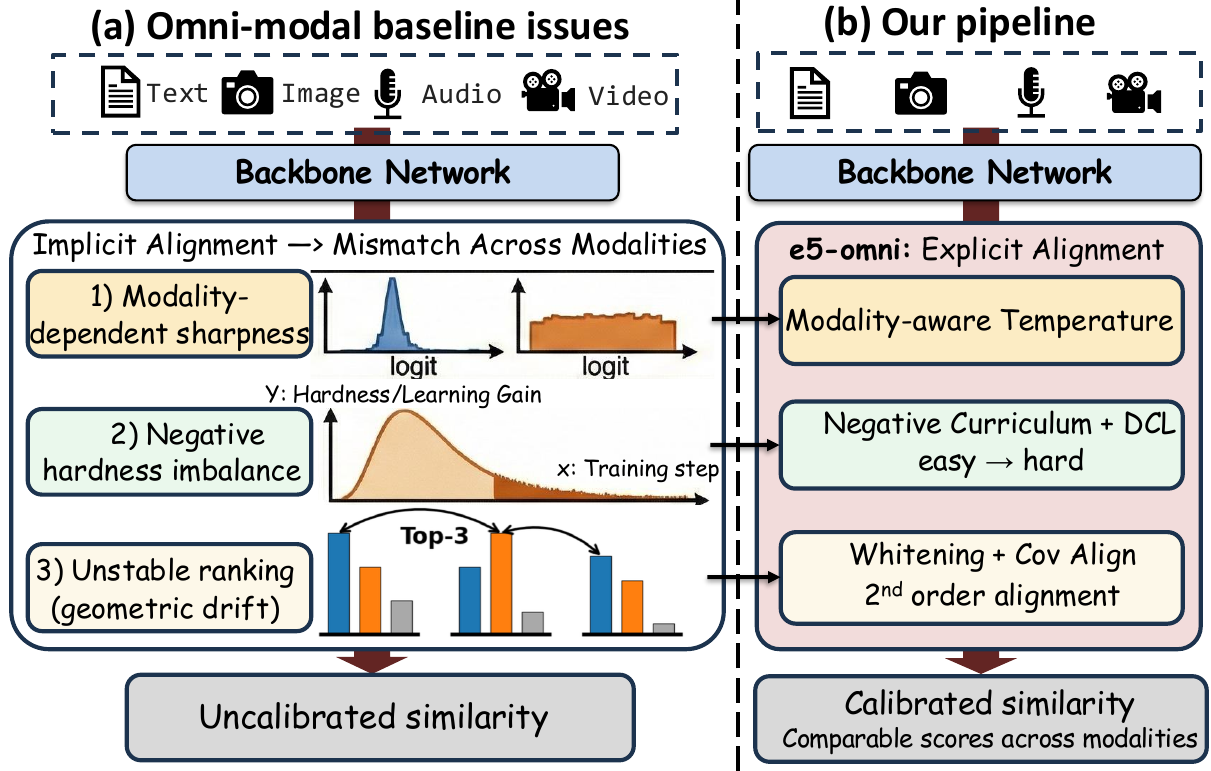}
	\caption{\textbf{Implicit vs.\ explicit alignment for omni-modal embeddings.}
(a) Implicit alignment leads to modality-dependent sharpness, negative hardness imbalance, and unstable ranking.
(b) \ours{} performs explicit alignment with three lightweight modules to calibrate cross-modality similarities.
}\label{fig:introduction}
\end{figure}

Recent work has started to move from bi-modal to omni-modal embedding models, e.g., by adopting a VLM-centered bi-encoder architecture~\citep{tevatron-omni,Nemotron,LCO-Embed}.
However, these methods still largely rely on implicit modality alignment inherited from the generative pretraining of VLMs. Contrastive fine-tuning provides only a limited corrective signal, and it may not fully calibrate modality-specific similarity scales or the local geometry of the shared space.
As shown in the left part of Fig.~\ref{fig:introduction}, omni-modal contrastive training is challenging because different modalities induce similarity scores with different scales, hardness profiles, and embedding geometry.
This often leads to three issues:
(1) Modality-dependent sharpness: using a single global temperature can produce overly sharp logits for some modality compositions but overly flat logits for others, which creates imbalanced contrastive gradients;
(2) Negative hardness imbalance: negative hardness varies across modalities and shifts during training; when all in-batch negatives are treated equally, optimization can be dominated by many trivially easy negatives, weakening the late-stage learning signal and limiting fine-grained discrimination in mixed-modality batches; and
(3) Unstable ranking: inconsistent geometry across heterogeneous inputs makes rankings sensitive to small score changes, even when all items are embedded in a single space.
These issues become more significant in the omni-modal setting, because mini-batches naturally mix diverse modalities and thus amplify scale and geometry mismatch.

In this work, we propose a lightweight explicit alignment method (\ours{}) that turns VLM backbones into robust omni-modal embedding models without modifying the architecture.
As shown in the right part of Fig.~\ref{fig:introduction}, \ours{} has three components:
(1) Modality-aware Temperature Calibration.
We introduce a lightweight calibration module that uses a trainable per-modality scaling vector to calibrate similarity logits across modalities.
This adaptive rescaling helps balance contrastive training signals across different modality compositions.
(2) Controllable Negative Curriculum.
We select in-batch negatives using a quantile-based threshold and gradually increase hardness after a warmup period.
To mitigate the bias introduced by negative selection, we further incorporate a debiased contrastive objective~\citep{dcl}, which stabilizes optimization in mixed-modality batches.
(3) Batch Whitening and Covariance Alignment.
We apply a batch whitening transform~\citep{Whiten} to the embeddings and add a CORAL-style regularizer~\citep{coral} to align mini-batch covariances across modalities.
This harmonizes second-order geometry in the shared space and leads to more consistent similarity comparisons across diverse omni-modal inputs.

Empirically, \ours{} delivers consistent improvements on diverse embedding benchmarks, including MMEB-V2~\citep{MMEB-v2} and the AudioCaps text--audio retrieval benchmark~\citep{AudioCaps}.
Moreover, the gains grow with model scale: upgrading the backbone from 3B to 7B yields larger improvements.
We also find that \ours{} transfers well to other VLM backbones.

\smallskip
In summary, our contributions are threefold:
\begin{itemize}
\item We propose \ours{}, a lightweight explicit alignment recipe that adapts generative VLMs into omni-modal embedding models without changing the backbone architecture.
\item We introduce three simple plug-in components: modality-aware logit calibration, a controllable negative curriculum with debiased contrastive learning, and batch whitening with covariance alignment, making omni-modal contrastive training more robust under mixed-modality batches.
\item We achieve consistent gains on MMEB-V2 and AudioCaps over strong baselines, and show that the recipe benefits more from larger backbones and transfers to other VLMs.
\end{itemize}

\section{Method: \ours{}}
\label{sec:method}

\begin{figure*}[t]
  \centering
  \includegraphics[width=0.98\textwidth]{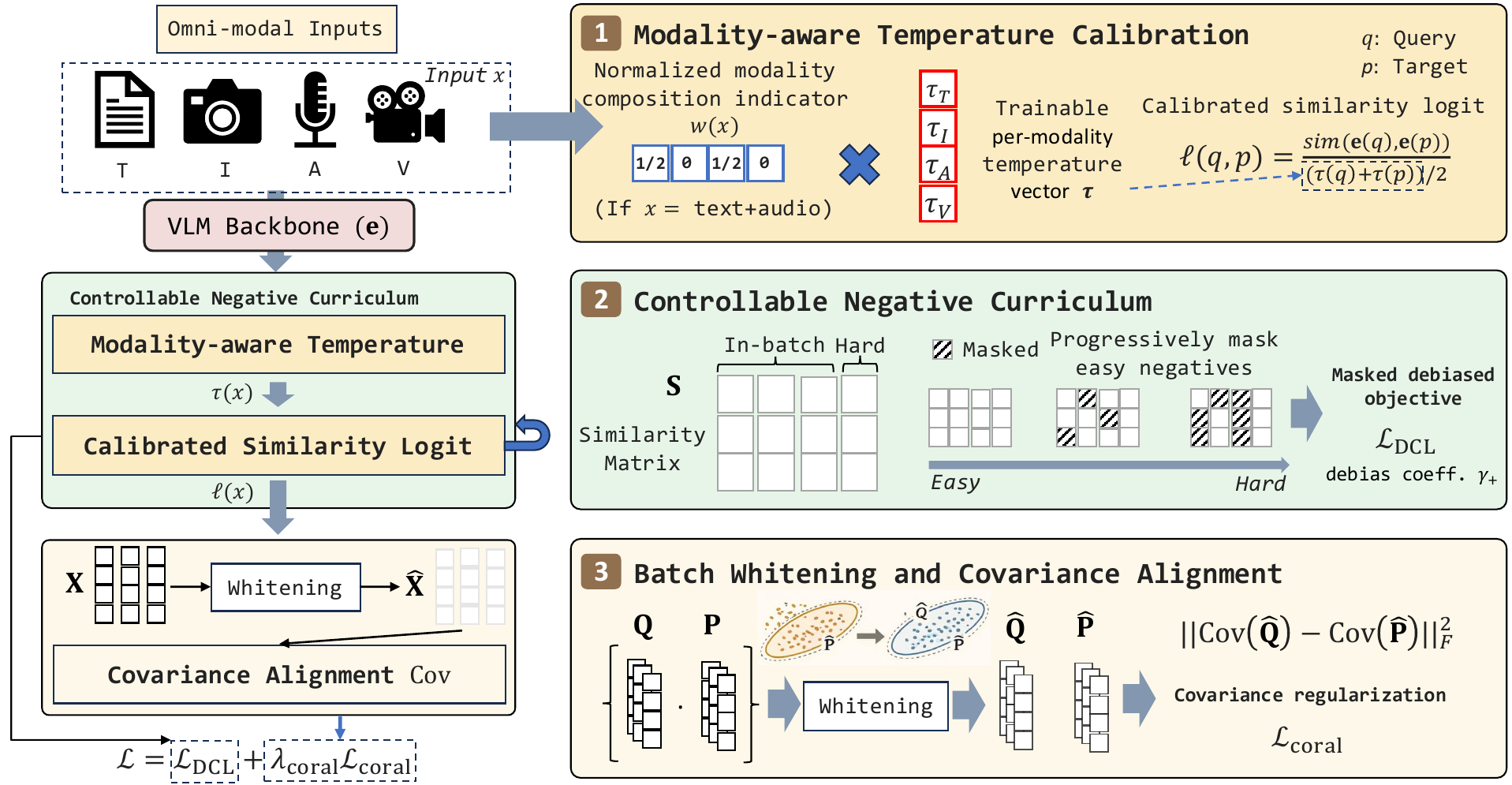}  
  \caption{\textbf{Overview of \ours{}.}
Given omni-modal inputs, \ours{} augments a VLM backbone with three lightweight components:
(1)~\textit{Modality-aware temperature calibration} computes a modality-composition indicator $w(x)$ and applies learnable modality-specific temperatures $\boldsymbol{\tau}$ to calibrate logits $\ell(q,p)$;
(2)~\textit{Controllable negative curriculum} progressively masks easy negatives and optimizes a masked debiased objective $\mathcal{L}_{\mathrm{DCL}}$;
(3)~\textit{Batch whitening and covariance alignment} whitens batch embeddings and adds a CORAL-style covariance regularizer.
}\label{fig:method_overview} 
\end{figure*}

In this section, we present \ours{}, a lightweight explicit-alignment framework that transforms a VLM into a unified omni-modal embedding model.
As shown in the left part of Fig.~\ref{fig:method_overview}, \ours{} preserves the backbone and introduces a simple training recipe.
Our framework comprises three components:
(1)~{Modality-aware temperature calibration}, which balances contrastive sharpness across modalities;
(2)~{A controllable negative curriculum}, which maintains a strong and stable learning signal; and
(3)~{Batch whitening and covariance alignment}, which harmonizes the shared-space geometry under heterogeneous inputs.
Together, these components improve the robustness of omni-modal embeddings without architectural modifications.

\subsection{Preliminaries}
We study omni-modal embedding with a base modality set
$\mathcal{M}_0=\{\texttt{T},\texttt{I},\texttt{A},\texttt{V}\}$,
corresponding to \emph{text}, \emph{image}, \emph{audio}, and \emph{video}.
Training data consist of tuples $(q, p^{+}, \mathcal{P}^{-})$,
where $q$ is a query, $p^{+}$ is its matched target, and $\mathcal{P}^{-}$ is a set of negative targets.
In this setting, $q$, $p^{+}$, and each $p^{-}\in\mathcal{P}^{-}$ can each take \emph{any} non-empty modality composition:
$
    q = x^{m_q}, p^{+}=x^{m_{+}}, p^{-}=x^{m_{-}}, \quad
    m_q,m_{+},m_{-}\in 2^{\mathcal{M}_0}\setminus\{\emptyset\}
$, where we denote an input with modality composition $m$ as $x^{m}$.

We consider two common sources of negatives: (i)~\textbf{in-batch negatives}, \eg, other positive targets in the same mini-batch, and
(ii)~\textbf{hard negatives}, mined by a retriever or constructed via heuristics.

Our objective is to learn a single embedding function $ \mathbf{e}(\cdot): \ \mathcal{X} \rightarrow \mathbb{R}^{D}, $
where $\mathcal{X}$ denotes the space of omni-modal inputs.
The embedding maps any omni-modal input into a shared representation space,
so that matched pairs $(q,p^{+})$ are close while mismatched pairs $(q,p^{-})$ are well separated.

\subsection{Modality-aware Temperature Calibration}
\label{subsec:method-modaltemp}

A single global temperature in contrastive learning implicitly assumes that all inputs induce similarity logits with comparable sharpness~\cite{simclr}.
In the omni-modal setting, different modalities can exhibit very different levels of ambiguity and noise.
As a result, a fixed temperature can make some modality compositions overly sharp while leaving others overly flat, leading to imbalanced gradients and unstable optimization.

To address this issue, we introduce a lightweight modality-aware temperature module implemented as a learnable per-modality scaling vector
$\boldsymbol{\tau}\in\mathbb{R}^{|\mathcal{M}_0|}$.
As shown in the top right part of Fig.~\ref{fig:method_overview}, for any input $x$, we construct a normalized modality-indicator weight
$w(x)\in\Delta^{|\mathcal{M}_0|-1}$ based on its modality composition.
Concretely, $w(x)$ assigns non-zero mass to every modality present in $x$ and is normalized by the number of active modalities.
We then define an \emph{instance temperature} as a weighted sum
$
\tau(x)=\max\!\big(w(x)^{\top}\boldsymbol{\tau},\,10^{-6}\big).
$
Given any pair $(q,p)$, we use a symmetric pairwise temperature
$
\tau(q,p)=\left(\tau(q)+\tau(p)\right)/2,
$
and compute the calibrated similarity logit as
\begin{equation}
\ell(q,p)=\frac{\mathrm{sim}(\mathbf{e}(q),\,\mathbf{e}(p))}{\tau(q,p)}\,.
\end{equation}
We use $\ell(\bullet,\bullet)$ consistently in the contrastive learning objective.

Intuitively, larger $\tau$ flattens logits while smaller $\tau$ sharpens logits, so noisier/ambiguous modalities tend to learn larger temperatures.
This stabilizes contrastive gradients in mixed omni-modal batches.

\subsection{Controllable Negative Curriculum}
\label{subsec:method-hns}
In omni-modal contrastive training, the hardness of negatives varies widely across batches and modalities, causing either weak supervision (overly easy negatives) or unstable optimization (false negatives).
To address this \emph{hardness imbalance}, we adopt a simple curriculum that gradually increases the focus on confusing negatives.
However, hard-negative selection is prone to \emph{selection bias}, since the hardest samples often include near-duplicates or semantically related items (\ie, false negatives), which can destabilize learning.
As shown in the middle right of Fig.~\ref{fig:method_overview}, we therefore combine negative selection with Debiased Contrastive Learning (DCL)~\cite{dcl}, which down-weights the aggregated negative to reduce the impact of potential false negatives.

For a mini-batch $\mathcal{B}=\{(q_i,p_i^{+},\mathcal{P}_i^{-})\}_{i=1}^{B}$, we form a similarity matrix
$\mathbf{S}\in\mathbb{R}^{B\times(B+K)}$, where the first $B$ columns correspond to the in-batch paired targets $\{p_j^{+}\}_{j=1}^{B}$, \ie, in-batch negatives
($S_{ij}=\ell(q_i,p_j^{+})$), and the remaining $K$ columns correspond to mined hard negatives
($S_{i,B+k}=\ell(q_i,p_{i,k}^{-})$). 
The diagonal $S_{ii}$ is the positive for row $i$.
We then mask negatives by keeping only the top fraction of the most confusing items per row.
Concretely, at training step $t$ we use a quantile threshold $\rho_t\in[0,1)$ and keep the hardest
$(1-\rho_t)$ fraction among all negatives in each row: for each $i$, let $\mathcal{N}_i(\rho_t)$ be the
set of indices of the largest $k=\lfloor(1-\rho_t)(B+K-1)\rfloor$ values in $\{S_{ij}\}_{j\neq i}$.
This produces a boolean mask $\mathbf{M}\in\{0,1\}^{B\times(B+K)}$ where $M_{ij}=1$ iff $j\in\mathcal{N}_i(\rho_t)$
(and $M_{ii}=0)$.

\paragraph{Masked DCL objective.}
We apply the mask inside a DCL loss by summing only over selected negatives.
Let $\Omega_i=\{i\}\cup \mathcal{N}_i(\rho_t)$.
DCL corrects the bias of (masked) negative sampling by subtracting a scaled positive term from the negative aggregate:
\begin{equation}
\mathcal{L}_{\mathrm{DCL}}
=-\frac{1}{B}\sum_{i=1}^{B}\log
\frac{\exp(S_{ii})}{\exp(S_{ii})+\widetilde{N}_i}\,,
\label{eq:dcl}
\end{equation}
$$
\widetilde{N}_i=\max\!\Big(\sum_{j\in\Omega_i\setminus\{i\}} \exp(S_{ij})-\gamma_{+}\exp(S_{ii}),\,\epsilon\Big)\,,
$$
where $\gamma_{+}\in(0,1)$ is the debiasing coefficient and $\epsilon$ is a small constant for numerical stability.
Here $\mathbf{S}$ is computed from the calibrated logit $\ell(\bullet,\bullet)$, hence no extra global temperature is required.

\paragraph{Curriculum schedule.}
Recall that $\rho_t$ denotes the \emph{mask ratio} of negatives, so a larger $\rho_t$ means fewer negatives are kept (and the remaining set is harder).
We therefore start with a smaller $\rho_t$ (keeping a larger fraction of negatives, including easier ones), and linearly increase it after a warmup period to gradually focus on fewer, harder negatives:
$$
\rho_t=\rho_{\mathrm{init}}+(\rho_{\mathrm{final}}-\rho_{\mathrm{init}})\,
\mathrm{clip}\!\left(\frac{t-t_0}{T-t_0},\,0,\,1\right),
$$
where $t$ is the training step, $t_0$ is the warmup cutoff (the step after which we start increasing $\rho_t$, i.e., hardness), $T$ is the total number of training steps, and $\mathrm{clip}(x,0,1)=\min(\max(x,0),1)$.

This curriculum lets the model first learn coarse omni-modal alignment with stable gradients, and then progressively resolve fine-grained confusions with genuinely hard negatives, while DCL mitigates the bias amplified by negative selection---together yielding stable optimization and cleaner alignment under mixed-modality batches.

\begin{table*}[t]
\centering
\caption{Overall results on MMEB-V2, which contains \textbf{78 tasks} spanning three major groups: \textbf{Image} (36 tasks), \textbf{Video} (18 tasks), and \textbf{VisDoc} (24 visual document tasks). 
We report group-wise averages and the overall score across all tasks.
We group prior baselines by their supported modality sets (\textit{bi-modal}, \textit{tri-modal}, and \textit{omni-modal}) and additionally report \ours{} w/o.\ explicit alignment (vanilla contrastive) as a strong baseline.
We highlight the best score in \textbf{bold} and the second-best with an \underline{underline}.
``$\dag$'' denotes our model outperforms all baselines significantly in paired t-test at $p<0.01$ level (with Bonferroni correction).
}\label{tab:mmeb-v2-overall}
\begin{tabular}{l c cccc}
\toprule
Model & Size & Image & Video & VisDoc & All \\
\midrule
\multicolumn{6}{l}{\textit{Text + Image (Bi-modal)}} \\
ColPali-v1.3~\cite{ColPali}        & 3B & 34.9 & 28.2   & 71.0   & 44.4 \\
VLM2Vec-v1~\cite{MMEB}       & 7B   & 65.5 & 33.7 & 46.4 & 52.3 \\
GME~\cite{GME}            & 7B   & 56.0 &	38.4 & {75.2} & 57.8 \\
CAFe~\cite{CAFe}      & 7B & 67.6 & 42.4   & 63.9   & 60.6 \\
\midrule
\multicolumn{6}{l}{\textit{Text + Image + Video}} \\
VLM2Vec-v2~\cite{MMEB-v2}       & 2B  & 64.9	& 34.6	& 65.4	& 58.0 \\
UniME-V2~\cite{UniME-V2}             & 7B  & {71.8}	& 39.0	& 56.7	& 59.6 \\
UME-R1~\cite{UME-R1}                 & 7B  & 71.3	& {47.5}	& 67.1	& \underline{64.5} \\
\midrule
\multicolumn{6}{l}{\textit{Text + Image + Video + Audio (Omni-modal)}} \\
Tevatron-Omni~\cite{tevatron-omni}   & 7B  & 37.1	& 35.1	& 74.5	& 48.1 \\
LCO-EMB~\cite{LCO-Embed}   & 7B  & 44.0 & 	38.2	& 69.8	& 50.6 \\
Omni-Embed-Nemotron~\cite{Nemotron}  & 3B  & 43.7	& 36.9	& 74.2	& 51.5 \\
\midrule
\multicolumn{6}{l}{\textit{Ours w/o. explicit alignment (vanilla contrastive)}} \\
\ours{}-vanilla-3B & 3B & 65.8 & 39.3 & 72.0 & 62.0 \\
\ours{}-vanilla-7B & 7B & 69.0 & 41.8 & 74.7 & 64.4 \\
\midrule
\multicolumn{6}{l}{\textit{Ours (Omni-modal)}} \\
\ours{}-3B & 3B & 67.6 & 40.6 & 73.2 & 63.1 \\
\ours{}-7B & 7B & 71.2 & 43.5 & 76.1 & \textbf{66.4}$^\dag$ \\
\bottomrule
\end{tabular}
\vspace{-0.12cm}
\end{table*}

\subsection{Batch Whitening and Covariance Alignment}
\label{subsec:method-align}

Representations produced by an omni-modal backbone often exhibit mismatched second-order statistics, which can distort similarity geometry and destabilize ranking~\cite{DBLP:journals/corr/abs-2103-15316}, especially when batches mix diverse modality compositions.
To explicitly regularize the shared embedding space, we apply batch whitening~\cite{Whiten} and add a CORAL-style covariance alignment loss~\cite{coral} as a regularizer.

For a mini-batch $\mathcal{B}=\{(q_i,p_i^{+},\mathcal{P}_i^{-})\}_{i=1}^{B}$, we compute embeddings with the shared encoder $\mathbf{e}(\cdot)$ and stack them as
$
\mathbf{Q}=[\mathbf{e}(q_i)]_{i=1}^{B}\in\mathbb{R}^{B\times D},
\mathbf{P}=[\mathbf{e}(p_i^{+})]_{i=1}^{B}\in\mathbb{R}^{B\times D}.
$
We compute a single whitening transform $W$ from the concatenation of query and target embeddings in the mini-batch, and apply the same transform to both sets:
$\widehat{\mathbf{Q}}=W(\mathbf{Q})$ and $\widehat{\mathbf{P}}=W(\mathbf{P})$.
We then define the empirical covariance operator for any batch embedding matrix $\mathbf{X}\in\mathbb{R}^{B\times D}$ (\eg, $\widehat{\mathbf{Q}}$ or $\widehat{\mathbf{P}}$) as
\begin{equation}
\mathrm{Cov}(\mathbf{X})=\frac{1}{B-1}(\mathbf{X}-\bar{\mathbf{X}})^{\top}(\mathbf{X}-\bar{\mathbf{X}})\,\,,
\label{eq:cov}
\end{equation}
where $\bar{\mathbf{X}}$ is the feature-wise mean over the batch.
The CORAL loss is
\begin{equation}
\mathcal{L}_{\mathrm{coral}}
=\frac{1}{4D^{2}}
\left\|
\mathrm{Cov}(\widehat{\mathbf{Q}})-\mathrm{Cov}(\widehat{\mathbf{P}})
\right\|_{F}^{2}.
\label{eq:coral_loss}
\end{equation}
With Eq.(\ref{eq:dcl}) and Eq.(\ref{eq:coral_loss}), the final objective is
\begin{equation}
\mathcal{L}
=\mathcal{L}_{\mathrm{DCL}}+\lambda_{\mathrm{coral}}\mathcal{L}_{\mathrm{coral}}\,\,,
\end{equation}
where $\lambda_{\mathrm{coral}}$ controls the strength of covariance regularization.
In practice, we implement $W(\cdot)$ with a numerically stable routine (\eg, group-wise computation with a small jitter) to avoid instability under high-dimensional embeddings.
By reducing second-order mismatch between $\widehat{\mathbf{Q}}$ and $\widehat{\mathbf{P}}$, this regularizer improves geometric consistency in the shared space and stabilizes similarity-based ranking in omni-modal retrieval.

\section{Experiments}

\subsection{Experimental Setup}
\label{subsec:setup}

\subsubsection{Training Datasets.}

We train \ours{} on a heterogeneous mixture of omni-modal retrieval pairs that cover diverse modality compositions.
In particular, the training data include:
(1) text-only contrastive pairs from BGE-m3~\cite{bge-m3};
(2) text--image pairs from the MMEB-V1 training set~\cite{MMEB} and PixMo-Docs~\cite{PixMo};
(3) text--video retrieval pairs from MSR-VTT~\cite{MSR-VTT} and the MMEB-V2 training set~\cite{MMEB-v2};
(4) text--audio retrieval pairs from AudioCaps~\cite{AudioCaps};
(5) visual-document retrieval pairs from MMEB-V2~\cite{MMEB-v2}.
Overall, this mixture exposes the model to queries/ targets that are uni-modal, bi-modal, or composed of multiple modalities, which is critical for learning a unified omni-modal embedding space.

\subsubsection{Implementation Details.}
\label{subsubsec:imple_dets}

We adopt Qwen2.5-Omni~\cite{Qwen2.5-Omni} as the VLM backbone and fine-tune it with LoRA~\cite{lora} for parameter-efficient adaptation.
We set the maximum sequence length to 512 tokens for both queries and targets, and train for one epoch with a learning rate of $1\times 10^{-4}$ and a warmup ratio of 0.005.
Training is performed on 8 H100 GPUs with a per-device batch size of 20 and gradient accumulation of 2. 
For each dataset, we use two dataset-provided hard negatives per query.

For our method-specific components, we initialize the trainable modality scaling vector $\boldsymbol{\tau}$ to 0.02.
After training, the learned temperatures are
$\boldsymbol{\tau}=[0.0130,\,0.0127,\,0.0219,\,0.0223]$
for $\{\texttt{T},\texttt{I},\texttt{A},\texttt{V}\}$, respectively.
We set the debiasing coefficient $\gamma_{+} = 0.1$, $\rho_{\mathrm{init}}=0.1$, $\rho_{\mathrm{final}}=0.5$, and use $t_0=4000$ warmup steps before increasing hardness.
For batch whitening and covariance alignment, we set $\lambda_{\mathrm{coral}}=0.05$.
We report the settings of \ours{}-7B.
More details are in Appendix~\ref{app:impl_more}.

\subsubsection{Evaluation.}

We evaluate \ours{} on two benchmarks.
First, we use \textbf{MMEB-V2}~\cite{MMEB-v2}, a large-scale multimodal embedding benchmark covering \emph{text, image, video, and visual documents}.
It contains \textbf{9 meta-tasks and 78 tasks} spanning diverse categories such as retrieval, classification, question answering, and visual document retrieval.
Following the MMEB-V2 protocol, we use Hit@1 as the primary metric for image/video tasks and report NDCG@5 for visual-document tasks.
Second, we evaluate audio retrieval on \textbf{AudioCaps}~\cite{AudioCaps}, which contains about 4.4K text--audio pairs.
We report Recall@1, \ie, the fraction of text queries whose matched audio is ranked at the top.

\begin{table}[t]
\centering
\small
\caption{Overall results on AudioCaps.
We only include \textit{omni-modal} embedding models that are trained with audio data.
``$\ddag$'' denotes our model outperforms all baselines significantly in paired t-test at $p<0.05$ level (with Bonferroni correction).
}\label{tab:audiocaps-overall}
\begin{tabular}{
p{0.52\linewidth}
>{\centering\arraybackslash}p{0.12\linewidth}
>{\centering\arraybackslash}p{0.20\linewidth}
}
\toprule
Model & Size & Recall@1 \\
\midrule
\multicolumn{3}{l}{\textit{\textbf{Baselines}}} \\
Tevatron-Omni              & 7B  & 34.0 \\
LCO-EMB                    & 7B  & 24.2 \\
Omni-Embed-Nemotron        & 3B  & 20.5 \\
\midrule
\multicolumn{3}{l}{\textit{\textbf{Ours}}} \\
\ours{}-3B & 3B & 34.3 \\
\ours{}-7B & 7B & \textbf{37.7}$^\ddag$ \\
\bottomrule
\end{tabular}
\end{table}

\subsection{Overall Results}

Table~\ref{tab:mmeb-v2-overall} reports overall performance on \textbf{MMEB-V2}, and Table~\ref{tab:audiocaps-overall} reports \textbf{AudioCaps} text--audio retrieval results.
\ours{} outperforms strong bi-modal and omni-modal baselines on both benchmarks, supporting the effectiveness of our framework.

We highlight three observations:
(1)~\textbf{Broad gains across modalities and tasks.}
On MMEB-V2, \ours{} improves image/video retrieval and also yields clear gains on the visual-document retrieval subset, suggesting that explicit alignment remains beneficial under heterogeneous, long-form visual inputs.
(2)~\textbf{Improved audio retrieval.}
On AudioCaps, \ours{}-7B achieves higher Recall@1 than omni-modal baselines, indicating that our alignment components generalize beyond vision-language and improve the stability of audio embeddings.
(3)~\textbf{Scaling with model size.}
The alignment gains grow when scaling from 3B to 7B (All: $+1.1$ vs.\ $+2.0$), suggesting that explicit alignment is complementary to model scaling.

\subsection{Ablation Study}
\label{sec:ablation}

\begin{table}[!t]
    \centering
    \small
    \caption{Ablation of \ours{}.
We remove one component at a time from the full model and report results on MMEB-V2 and AudioCaps.
}\label{tab:ablation}
    \begin{tabular}{
p{0.5\linewidth} @{}cc
} 
    \toprule
    Model & MMEB-V2 & AudioCaps \\
    \midrule
    \ours{}-7B & \textbf{66.4} & \textbf{37.7} \\
    \midrule
    \quad w/o. Modality-aware Temp. & 65.7 & 36.6 \\
    \quad w/o. Curriculum Schedule & 65.7 & 36.7 \\
    \quad w/o. DCL (w/ Curriculum)  & 66.1 & 37.0 \\
    \quad w/o. Whitening \& CORAL   & 65.9 & 36.3 \\
    \bottomrule
    \end{tabular}
\end{table}

To understand the contribution of each component in \ours{}, we conduct ablation studies by removing one design choice at a time while keeping all other settings fixed.
We report results on MMEB-V2 and AudioCaps in Table~\ref{tab:ablation}.
Overall, removing any component degrades performance, indicating that the proposed techniques are complementary.

\noindent\textbf{Modality-aware Temperature Calibration.}
Disabling modality-aware temperature calibration (\ie, using a single global temperature of 0.02, matching the initialization of $\boldsymbol{\tau}$) causes a clear performance drop.
This suggests that calibrating modality-dependent logit sharpness stabilizes contrastive optimization under mixed omni-modal batches.

\noindent\textbf{Controllable Negative Curriculum.}
We ablate the negative curriculum by (i) removing the curriculum schedule and using a fixed threshold 0.3, or (ii) removing DCL while keeping curriculum.
Both variants underperform \ours{}, indicating that progressively increasing negative hardness helps avoid unstable early training while improving fine-grained discrimination later.
We further observe additional degradation when removing DCL, suggesting that debiasing is important when hard-negative selection amplifies false-negative bias.

\noindent\textbf{Batch Whitening and Covariance Alignment.}
Removing whitening and the covariance regularizer also reduces performance on both benchmarks.
This supports our motivation that regularizing second-order statistics improves geometric consistency in the shared embedding space and stabilizes similarity-based ranking.

\subsection{Embedding-space Modality Alignment Diagnostics}
\label{subsec:exp-align-viz}

Retrieval metrics summarize ranking quality but do not reveal whether the embedding space is distributionally consistent across modalities.
We therefore run a lightweight distribution-level diagnostic on VOC2007, a clean text--image setting for inspecting cross-modal embedding mismatch.
We compare \ours{} with \ours{}~w/o.\ alignment (7B), which disables all three techniques in Sec.~\ref{sec:method} while keeping the same backbone and embedding dimension.

\paragraph{Setup.}
We sample $1$k text queries and $2$k image targets, extract embeddings with $\mathbf{e}(\cdot)$, and stack them as $\mathbf{Q}\in\mathbb{R}^{N_q\times D}$ and $\mathbf{P}\in\mathbb{R}^{N_p\times D}$.
Image targets are drawn from the intersection of candidate pools for both checkpoints.
We compute empirical means and $\mathrm{Cov}(\cdot)$ as in Eq.~(\ref{eq:cov}).

\begin{figure}[t]
  \centering
  \includegraphics[width=0.98\columnwidth]{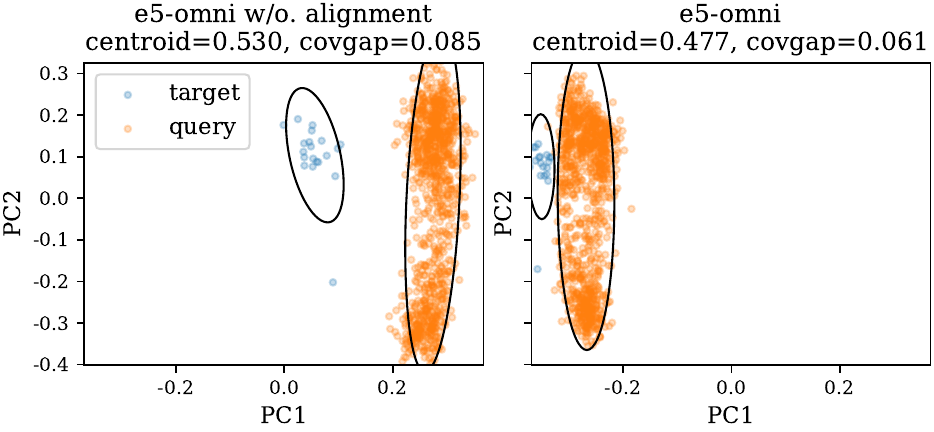}
  \caption{{PCA overlap on VOC2007.}
Left: \ours{} w/o.\ alignment. Right: \ours{}.
We project embeddings into a shared 2D PCA space and overlay $2\sigma$ covariance ellipses.
We report \emph{centroid} (distance between the query/target mean embeddings) and \emph{covgap} (Frobenius gap between their covariance matrices).}\label{fig:align-pca-voc}
\end{figure}

\begin{figure}[t]
  \centering
  \includegraphics[width=0.98\columnwidth]{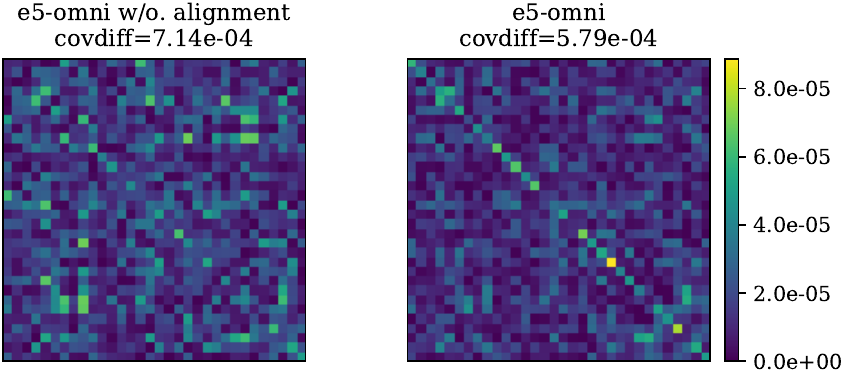}
  \caption{{Covariance-difference heatmap on VOC2007.}
Left: \ours{} w/o.\ alignment. Right: \ours{}.
After a fixed 32D random projection, we visualize \emph{covdiff}: the entrywise magnitude of the query--target covariance difference matrix.}\label{fig:align-covdiff-voc}
\end{figure}

\paragraph{Diagnostics.}
\emph{PCA overlap.}
We fit a single PCA basis on the union of $\mathbf{Q}$ and $\mathbf{P}$ from both models, project all embeddings to 2D, and visualize the query/target clouds with $2\sigma$ covariance ellipses (Fig.~\ref{fig:align-pca-voc}).
We quantify distribution mismatch using the centroid gap $\lVert \mu_{\mathbf{Q}}-\mu_{\mathbf{P}}\rVert_2$ and the covariance gap $\lVert \mathrm{Cov}(\mathbf{Q})-\mathrm{Cov}(\mathbf{P})\rVert_F$.
\emph{Covariance-difference heatmap.}
To inspect second-order mismatch more directly, we apply a fixed random Gaussian projection to 32D and visualize the entrywise magnitude (Fig.~\ref{fig:align-covdiff-voc}), together with its Frobenius norm.

\paragraph{Results and interpretation.}
Compared to \ours{}~w/o.\ alignment, the full \ours{} yields smaller query--target mismatch in embedding space: the centroid gap and the covariance gap decrease in Fig.~\ref{fig:align-pca-voc}.
The covariance-difference heatmap in Fig.~\ref{fig:align-covdiff-voc} further indicates reduced second-order discrepancy with fewer high-magnitude entries.
While the heatmap mainly reflects second-order geometry (most directly tied to whitening/covariance regularization), the PCA overlap reflects both first- and second-order effects and may also be influenced by training dynamics (\eg, modality-aware logit calibration and more stable negative supervision).
Overall, these diagnostics provide distribution-level evidence that the \emph{combined} design of \ours{} promotes a more consistent shared embedding space, complementing the benchmark results.

\subsection{Hyperparameter Analysis}
\label{subsec:param}

\begin{figure}[t]
\centering
\includegraphics[width=0.5\textwidth]{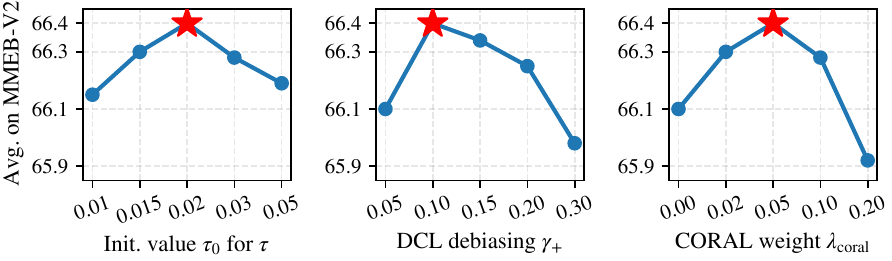}
\caption{The performances of \ours{}-7B under different training settings on MMEB-V2. We report the overall scores using same metric as in Table~\ref{tab:mmeb-v2-overall}.}\label{fig:param} 
\end{figure}

We analyze the sensitivity of \ours{} to key hyperparameters in our explicit alignment recipe.
We tune these hyperparameters on 1K-sample validation splits drawn from the corresponding training sets.
For consistency with prior experiments, we report results on the MMEB-V2 test sets in Fig.~\ref{fig:param}.
Additional studies are deferred to Appendix~\ref{app:param}.

\noindent\textbf{Modality-aware temperature initialization.}
Our modality-aware calibration introduces a trainable per-modality scaling vector
$\boldsymbol{\tau}$ (Sec.~\ref{subsec:method-modaltemp}).
We sweep the initialization value $\tau_0$ (\ie, initializing all entries of $\boldsymbol{\tau}$ to the same constant $\tau_0$).
We observe a trade-off: overly small $\tau_0$ yields overly sharp logits early in training, while large $\tau_0$ produces overly flat logits and slows convergence.
Overall, initializing $\boldsymbol{\tau}$ in a small range (\eg, $\tau_0\in[0.015,0.03]$) is robust, and we use $\tau_0=0.02$ by default.

\noindent\textbf{DCL debiasing coefficient.}
We sweep the DCL parameter $\gamma_{+}$ (Sec.~\ref{subsec:method-hns}), which controls the strength of debiasing under negative selection.
Moderate values work best: too small $\gamma_{+}$ provides limited correction, while too large $\gamma_{+}$ can over-correct and weaken supervision.
In our experiments, $\gamma_{+}\in[0.1,0.2]$ yields consistent improvements on MMEB-V2, and we use $\gamma_{+}=0.1$ by default.

\noindent\textbf{Covariance regularization weight.}
Finally, we vary the CORAL weight $\lambda_{\mathrm{coral}}$ (Sec.~\ref{subsec:method-align}).
Increasing $\lambda_{\mathrm{coral}}$ generally improves cross-modal geometric consistency, but overly large weights can interfere with the retrieval objective.
We observe that a small auxiliary weight (\eg, $\lambda_{\mathrm{coral}}\in[0.02,0.1]$) provides the best trade-off.

\subsection{Generalization to Other VLM Backbones}
\label{subsec:generalization}

To validate that \ours{} is not tied to a specific backbone, we apply our explicit alignment recipe to multiple off-the-shelf VLMs with different sizes and architectures.
For each backbone, we first train a strong baseline w/o.\ alignment (as defined in Sec.~\ref{subsec:exp-align-viz}), and then train the full \ours{} by enabling all three techniques.

We report results on MMEB-V2 in Table~\ref{tab:generalization}.
We omit AudioCaps because only a limited number of public VLM backbones support audio inputs out of the box, making a broad backbone comparison infeasible.
Overall, \ours{} consistently improves over the corresponding baselines across all backbones, demonstrating that our method is a plug-and-play strategy for turning diverse VLM backbones into robust omni-modal embedding models.

\section{Related Work}

\noindent \textbf{Bi-modal Embedding Models.}
Early vision--language embedding models primarily focus on aligning text and images in a shared space, typically via dual-encoder contrastive learning~\citep{ALIGN,CLIP,BLIP,blip2}.
For example, CLIP~\citep{CLIP} learns image and text encoders by matching paired image--caption data at scale.
More recently, a line of work builds retrieval-oriented bi-encoders on top of modern VLMs~\citep{MMEB,GME,mme5,B3,Moca}, converting generative or instruction-following backbones into embedding models.
For example, mmE5~\citep{mme5} improves text--image embedding quality through data synthesis.
Despite strong progress on text--image representations, these bi-modal models are not designed to fully meet modern retrieval needs that increasingly involve additional modalities.

\noindent \textbf{Omni-modal Embedding Models.}
Moving beyond text--image, recent work has started to incorporate video~\citep{MMEB-v2,UME-R1,UniME-V2}.
For example, UME-R1~\citep{UME-R1} explores a reasoning-driven generative embedding paradigm that leverages reinforcement learning.
More recent efforts further expand to video and audio modalities~\citep{tevatron-omni,Nemotron,LCO-Embed}.
For example, Omni-Embed-Nemotron~\citep{Nemotron} builds a unified embedding model on a Qwen-Omni backbone to support omni-modal retrieval in a single shared space.
Similarly, LCO-Embed~\citep{LCO-Embed} proposes a language-centric omni-modal embedding framework.

While these omni-modal models broaden coverage, they largely rely on VLM pretraining to provide implicit alignment and typically lack explicit calibration and alignment mechanisms.
This limitation becomes more critical as the number of supported modalities grows.
Unlike these models, \ours{} introduces a lightweight, plug-and-play explicit alignment recipe to improve robustness under mixed-modality training.

\begin{table}[t]
    \centering
    \small
    \caption{Generalization of \ours{} across different VLM backbones on MMEB-V2. ``w/o. alignment'' disables all alignment components.}\label{tab:generalization}
    \begin{tabular}{l c c l
    }
    \toprule
    \textbf{VLM Backbone} & \textbf{Size} & \textbf{Method} & \textbf{MMEB-V2} \\
    \midrule
    \multirow{2}{*}{Qwen2.5-VL} & \multirow{2}{*}{3B} 
        & w/o. alignment & 61.5 \\
     &  & \ours{} & 62.9 $\uparrow$ \\
    \midrule
    \multirow{2}{*}{Qwen2-VL} & \multirow{2}{*}{2B} 
        & w/o. alignment & 59.2 \\
     &  & \ours{} & 60.5 $\uparrow$ \\
    \midrule
    \multirow{2}{*}{LLaVA-OV} & \multirow{2}{*}{7B} 
        & w/o. alignment & 63.7 \\
     &  & \ours{} & 65.4 $\uparrow$ \\
    \bottomrule
    \end{tabular}
\end{table}

\section{Conclusion}

We introduced \ours{}, a lightweight explicit-alignment recipe that turns off-the-shelf VLM backbones into unified omni-modal embedding models.
\ours{} keeps the backbone unchanged and adds three simple components: modality-aware temperature calibration to balance logit sharpness across modality compositions; a controllable negative curriculum (with debiasing) to stabilize contrastive optimization under mixed-modality batches; and batch whitening with covariance alignment to harmonize second-order geometry in the shared space.
Experiments on MMEB-V2 and AudioCaps show consistent gains over strong baselines, indicating that explicit alignment yields more robust omni-modal embeddings with minimal overhead.

\clearpage

\section*{Limitations}
While \ours{} achieves strong omni-modal embedding performance with minimal overhead, several limitations remain:

\begin{enumerate}
    \item \textbf{Scope of alignment.}
    Our recipe primarily targets similarity geometry and optimization dynamics for retrieval.
    It does not explicitly improve higher-level reasoning or compositional understanding, so gains may be smaller on tasks that require multi-step inference beyond embedding similarity.

    \item \textbf{Dependence on batch statistics.}
    Batch whitening and covariance alignment rely on mini-batch estimates, whose quality can vary with batch size, modality composition, and distributed training settings.
    Although we apply stabilizing tricks (\eg, jitter and shrinkage), these estimates may still be noisy for small or highly imbalanced omni-modal batches.

    \item \textbf{Training and evaluation coverage.}
    We evaluate on MMEB-V2 and AudioCaps, which cover text/image/video and text--audio retrieval.
    Broader evaluation on more diverse audio/video domains, long-horizon retrieval settings, and real-world multimodal corpora would further strengthen the conclusions.
\end{enumerate}


\appendix

\clearpage

\section*{Appendix}

\section{Detailed Results on MMEB-V2}

We report detailed results of \ours{} and strong baselines on MMEB-V2, which contains 78 tasks spanning three modality groups: \textbf{Image}, \textbf{Video}, and \textbf{Visual Document} (VisDoc).
Table~\ref{tab:full_score_image} summarizes overall performance and provides a breakdown of the \textbf{Image} group (36 tasks; Hit@1), including both meta-task averages and per-dataset results.
Table~\ref{tab:full_score_video} presents detailed \textbf{Video} results (18 tasks; Hit@1).
Table~\ref{tab:full_score_visdoc} reports detailed \textbf{VisDoc} results (24 tasks; NDCG@5), covering document-level retrieval benchmarks such as ViDoRe.
For readability and to fit the appendix layout, we split the detailed results into three tables.

\definecolor{avgcolor}{RGB}{242,248,236}     
\definecolor{icatcolor}{RGB}{255,243,233}    
\definecolor{vcatcolor}{RGB}{235,248,255}    
\definecolor{vdcatcolor}{RGB}{250,236,244}   

\begin{table*}[!t]
    \centering
\caption{Detailed results on the full MMEB-v2 benchmark: summary and image tasks (36 tasks, Hit@1).}
\label{tab:full_score_image}
    \renewcommand{\arraystretch}{1.1}
    \begin{adjustbox}{width=\textwidth}
    \begin{tabular}{l|ccccccccc}
        \toprule
        ~ & GME & VLM2Vec-v1 & VLM2Vec-v2 & CAFe & UME-R1 & LCO-EMB & Omni-Embed-Nemotron & \ours{} & \ours{} \\
        Model Size & 7B & 7B & 2B & 7B & 7B & 7B & 3B & 3B & 7B \\
        \midrule
        \rowcolor{avgcolor}
        Avg - All (78 tasks) & 57.8 & 52.3 & 58.0 & 60.6 & 64.5 & 50.6 & 51.5 & 63.1 & 66.4 \\
        \rowcolor{avgcolor}
        Avg - Image (36 tasks, Hit@1) & 56.0 & 65.5 & 64.9 & 67.6 & 71.3 & 44.0 & 43.7 & 67.6 & 71.2 \\
        \rowcolor{avgcolor}
        Avg - Video (18 tasks, Hit@1) & 38.4 & 33.7 & 34.6 & 42.4 & 47.5 & 38.2 & 36.9 & 40.6 & 43.5 \\
        \rowcolor{avgcolor}
        Avg - Visdoc (24 tasks, NDCG@5) & 75.2 & 46.4 & 65.4 & 63.9 & 67.1 & 69.8 & 74.2 & 73.2 & 76.1 \\
        \rowcolor{icatcolor}
        I-CLS (10) & 57.7 & 62.7 & 62.9 & 63.6 & 67.1 & 56.3 & 46.0 & 64.7 & 66.7 \\
        \rowcolor{icatcolor}
        I-QA (10) & 34.7 & 56.9 & 56.3 & 61.7 & 69.2 & 16.9 & 20.5 & 62.1 & 68.5 \\
        \rowcolor{icatcolor}
        I-RET (12) & 71.2 & 69.4 & 69.5 & 69.1 & 71.9 & 52.2 & 58.0 & 70.5 & 73.0 \\
        \rowcolor{icatcolor}
        I-VG (4) & 59.3 & 82.2 & 77.3 & 87.6 & 84.9 & 57.0 & 52.9 & 79.6 & 83.9 \\
        \rowcolor{vcatcolor}
        V-CLS (5) & 37.4 & 39.1 & 39.3 & 35.8 & 48.6 & 39.3 & 40.5 & 40.2 & 46.6 \\
        \rowcolor{vcatcolor}
        V-QA (5) & 50.4 & 30.0 & 34.3 & 58.7 & 60.7 & 57.6 & 44.3 & 48.5 & 52.9 \\
        \rowcolor{vcatcolor}
        V-RET (5) & 28.4 & 29.0 & 28.8 & 34.4 & 38.2 & 24.8 & 32.7 & 33.1 & 36.7 \\
        \rowcolor{vcatcolor}
        V-MR (3) & 37.0 & 38.9 & 36.8 & 39.5 & 39.3 & 26.5 & 25.6 & 40.7 & 34.2 \\
        \rowcolor{vdcatcolor}
        VD-Vidore-V1 (10) & 89.4 & 56.9 & 75.5 & 70.7 & 75.7 & 80.4 & 85.7 & 83.1 & 87.6 \\
        \rowcolor{vdcatcolor}
        VD-Vidore-V2 (4) & 55.6 & 9.4 & 44.9 & 49.6 & 50.5 & 56.4 & 61.3 & 59.5 & 62.4 \\
        \rowcolor{vdcatcolor}
        VD-VisRAG (6) & 85.0 & 59.1 & 79.4 & 79.5 & 83.7 & 79.7 & 84.3 & 85.5 & 87.5 \\
        \rowcolor{vdcatcolor}
        VD-OOD (4) & 44.4 & 38.1 & 39.4 & 38.1 & 37.6 & 41.8 & 43.3 & 43.4 & 44.3 \\
        \midrule
        ImageNet-1K & 64.6 & 80.1 & 80.8 & 77.3 & 80.4 & 73.0 & 57.0 & 76.5 & 79.4 \\
        N24News & 50.5 & 79.7 & 72.9 & 83.2 & 82.3 & 37.5 & 36.4 & 76.7 & 77.2 \\
        HatefulMemes & 53.6 & 69.7 & 56.3 & 78.7 & 79.0 & 45.1 & 48.0 & 66.2 & 72.3 \\
        VOC2007 & 80.3 & 80.7 & 85.0 & 89.8 & 90.8 & 64.9 & 47.9 & 84.8 & 83.9 \\
        SUN397 & 69.5 & 77.4 & 71.0 & 79.9 & 80.3 & 72.0 & 53.1 & 75.1 & 76.1 \\
        Place365 & 39.1 & 37.4 & 35.9 & 45.0 & 46.8 & 40.8 & 30.7 & 41.7 & 44.3 \\
        ImageNet-A & 41.2 & 58.1 & 47.4 & 55.2 & 53.9 & 52.2 & 37.0 & 58.6 & 57.2 \\
        ImageNet-R & 83.9 & 73.9 & 89.3 & 88.0 & 90.1 & 88.9 & 80.9 & 85.5 & 84.5 \\
        ObjectNet & 69.0 & 40.1 & 65.2 & 22.5 & 42.3 & 61.7 & 44.9 & 53.0 & 57.7 \\
        Country211 & 24.8 & 29.8 & 25.2 & 16.7 & 25.0 & 26.5 & 24.1 & 29.1 & 34.7 \\
        OK-VQA & 33.2 & 56.8 & 51.5 & 67.3 & 71.7 & 14.4 & 18.6 & 64.9 & 70.5 \\
        A-OKVQA & 21.0 & 47.3 & 43.6 & 63.8 & 58.7 & 10.9 & 11.9 & 54.8 & 59.7 \\
        DocVQA & 41.4 & 89.7 & 90.1 & 79.2 & 93.8 & 17.5 & 17.3 & 91.2 & 93.3 \\
        InfographicsVQA & 20.3 & 60.0 & 58.8 & 53.3 & 79.2 & 7.4 & 9.1 & 52.8 & 62.9 \\
        ChartQA & 17.8 & 56.9 & 47.4 & 48.8 & 75.1 & 12.7 & 13.1 & 52.7 & 62.8 \\
        Visual7W & 22.2 & 52.7 & 52.9 & 52.5 & 55.2 & 10.8 & 7.4 & 59.8 & 65.2 \\
        ScienceQA & 28.0 & 38.5 & 38.2 & 65.4 & 53.7 & 26.7 & 25.3 & 43.6 & 56.4 \\
        VizWiz & 39.0 & 39.9 & 43.3 & 43.8 & 51.6 & 31.8 & 34.8 & 53.1 & 56.3 \\
        GQA & 76.9 & 55.1 & 64.9 & 65.7 & 69.3 & 11.5 & 31.2 & 65.4 & 72.1 \\
        TextVQA & 46.8 & 71.6 & 72.2 & 76.8 & 83.5 & 25.1 & 36.6 & 82.8 & 86.2 \\
        VisDial & 60.8 & 81.9 & 82.7 & 82.7 & 80.7 & 51.1 & 51.5 & 78.9 & 82.4 \\
        CIRR & 54.9 & 51.1 & 57.5 & 60.4 & 55.3 & 8.4 & 14.8 & 54.9 & 54.9 \\
        VisualNews\_t2i & 79.7 & 80.5 & 74.5 & 69.5 & 76.8 & 61.4 & 58.8 & 74.6 & 75.3 \\
        VisualNews\_i2t & 83.6 & 81.2 & 78.2 & 79.4 & 82.0 & 59.5 & 56.3 & 77.7 & 79.8 \\
        MSCOCO\_t2i & 71.2 & 77.2 & 75.3 & 75.4 & 78.3 & 66.1 & 60.7 & 78.1 & 76.6 \\
        MSCOCO\_i2t & 57.7 & 73.9 & 71.4 & 73.1 & 71.4 & 59.5 & 56.5 & 72.5 & 76.9 \\
        NIGHTS & 67.6 & 67.6 & 68.6 & 66.7 & 68.1 & 65.2 & 65.3 & 68.4 & 68.6 \\
        WebQA & 91.4 & 88.3 & 90.6 & 89.3 & 90.9 & 53.2 & 89.8 & 89.2 & 89.3 \\
        FashionIQ & 37.8 & 17.1 & 19.5 & 39.0 & 23.4 & 3.5 & 6.2 & 19.5 & 24.5 \\
        Wiki-SS-NQ & 78.2 & 62.3 & 66.9 & 61.2 & 72.5 & 64.2 & 87.7 & 74.9 & 79.4 \\
        OVEN & 75.1 & 66.5 & 64.3 & 60.8 & 71.4 & 59.9 & 67.9 & 76.1 & 78.7 \\
        EDIS & 96.0 & 85.7 & 84.1 & 71.3 & 92.0 & 74.4 & 80.5 & 81.7 & 89.6 \\
        MSCOCO & 31.4 & 75.7 & 67.1 & 84.7 & 72.7 & 33.8 & 32.8 & 70.9 & 74.8 \\
        RefCOCO & 60.9 & 87.6 & 87.1 & 89.4 & 91.4 & 50.9 & 59.7 & 85.3 & 89.4 \\
        RefCOCO-Matching & 78.4 & 84.6 & 85.8 & 83.0 & 91.1 & 98.5 & 65.9 & 75.8 & 87.2 \\
        Visual7W-Pointing & 66.5 & 81.0 & 69.2 & 93.2 & 84.2 & 44.6 & 53.1 & 86.5 & 84.3 \\
        \bottomrule
    \end{tabular}
    \end{adjustbox}
\end{table*}

\begin{table*}[!t]
    \centering
\caption{Detailed results on the full MMEB-v2 benchmark: video tasks (18 tasks, Hit@1).}
\label{tab:full_score_video}
    \renewcommand{\arraystretch}{1.1}
    \begin{adjustbox}{width=\textwidth}
    \begin{tabular}{l|ccccccccc}
        \toprule
        ~ & GME & VLM2Vec-v1 & VLM2Vec-v2 & CAFe & UME-R1 & LCO-EMB & Omni-Embed-Nemotron & \ours{} & \ours{} \\
        Model Size & 7B & 7B & 2B & 7B & 7B & 7B & 3B & 3B & 7B \\
        \midrule
        K700 & 39.7 & 35.5 & 38.0 & 40.1 & 42.8 & 40.2 & 36.0 & 41.9 & 48.0 \\
        SmthSmthV2 & 30.6 & 32.1 & 42.8 & 35.8 & 50.4 & 33.3 & 33.4 & 34.4 & 41.3 \\
        HMDB51 & 47.9 & 42.2 & 40.9 & 46.9 & 58.3 & 37.3 & 49.1 & 42.2 & 54.4 \\
        UCF101 & 54.7 & 61.8 & 60.0 & 39.6 & 70.0 & 66.0 & 60.5 & 58.0 & 66.0 \\
        Breakfast & 14.3 & 23.8 & 14.8 & 16.6 & 21.5 & 19.9 & 23.3 & 24.5 & 23.3 \\
        MVBench & 46.6 & 28.5 & 33.7 & 48.9 & 58.2 & 58.1 & 40.6 & 45.3 & 50.3 \\
        Video-MME & 39.2 & 27.8 & 30.7 & 46.0 & 47.3 & 48.5 & 38.6 & 43.6 & 45.6 \\
        NExTQA & 53.6 & 20.3 & 20.9 & 62.4 & 69.6 & 65.5 & 41.2 & 54.4 & 62.1 \\
        EgoSchema & 46.8 & 21.8 & 34.0 & 60.0 & 52.4 & 40.4 & 37.6 & 36.6 & 39.0 \\
        ActivityNetQA & 65.6 & 51.4 & 52.3 & 76.0 & 76.0 & 75.3 & 63.4 & 62.8 & 67.7 \\
        DiDeMo & 26.4 & 29.3 & 30.4 & 37.8 & 40.0 & 26.9 & 34.7 & 34.8 & 36.1 \\
        MSR-VTT & 31.8 & 34.5 & 28.3 & 36.5 & 38.9 & 22.0 & 35.4 & 36.0 & 42.1 \\
        MSVD & 49.7 & 46.7 & 48.1 & 56.4 & 60.8 & 50.7 & 54.0 & 52.7 & 57.2 \\
        VATEX & 24.9 & 25.5 & 26.5 & 32.0 & 32.6 & 18.8 & 25.7 & 28.6 & 33.5 \\
        YouCook2 & 9.1 & 9.0 & 10.6 & 9.5 & 18.5 & 5.4 & 13.8 & 13.7 & 14.5 \\
        QVHighlight & 59.5 & 57.7 & 49.4 & 58.4 & 54.9 & 20.0 & 29.1 & 62.6 & 42.4 \\
        Charades-STA & 14.0 & 19.8 & 20.2 & 18.7 & 21.9 & 30.0 & 11.6 & 18.2 & 17.9 \\
        MomentSeeker & 37.4 & 39.3 & 40.8 & 41.4 & 41.1 & 29.5 & 36.1 & 41.3 & 42.3 \\
        \bottomrule
    \end{tabular}
    \end{adjustbox}
\end{table*}

\begin{table*}[!t]
    \centering
\caption{Detailed results on the full MMEB-v2 benchmark: VisDoc tasks (24 tasks, NDCG@5).}
\label{tab:full_score_visdoc}
    \renewcommand{\arraystretch}{1.1}
    \begin{adjustbox}{width=\textwidth}
    \begin{tabular}{l|ccccccccc}
        \toprule
        ~ & GME & VLM2Vec-v1 & VLM2Vec-v2 & CAFe & UME-R1 & LCO-EMB & Omni-Embed-Nemotron & \ours{} & \ours{} \\
        Model Size & 7B & 7B & 2B & 7B & 7B & 7B & 3B & 3B & 7B \\
        \midrule
        ViDoRe\_arxivqa & 86.9 & 60.2 & 80.6 & 73.3 & 73.6 & 79.1 & 85.3 & 82.2 & 87.6 \\
        ViDoRe\_docvqa & 57.5 & 34.7 & 44.9 & 38.3 & 41.1 & 50.8 & 57.8 & 52.0 & 58.0 \\
        ViDoRe\_infovqa & 91.6 & 70.4 & 83.7 & 80.6 & 80.8 & 84.5 & 89.8 & 88.9 & 92.3 \\
        ViDoRe\_tabfquad & 94.6 & 78.2 & 89.2 & 80.7 & 90.2 & 92.0 & 91.7 & 88.9 & 93.2 \\
        ViDoRe\_tatdqa & 74.1 & 27.6 & 43.8 & 37.8 & 46.7 & 58.7 & 70.2 & 58.8 & 70.9 \\
        ViDoRe\_shiftproject & 96.8 & 38.6 & 60.8 & 52.0 & 65.0 & 72.2 & 80.2 & 80.1 & 85.8 \\
        ViDoRe\_artificial\_intelligence & 99.6 & 67.7 & 88.5 & 86.0 & 89.5 & 92.6 & 97.0 & 98.2 & 98.2 \\
        ViDoRe\_energy & 95.3 & 60.4 & 86.5 & 84.8 & 85.7 & 88.1 & 92.6 & 93.2 & 95.0 \\
        ViDoRe\_government\_reports & 98.8 & 61.8 & 85.0 & 85.0 & 89.8 & 91.6 & 95.6 & 93.2 & 96.3 \\
        ViDoRe\_healthcare\_industry & 99.3 & 69.9 & 92.2 & 88.4 & 94.3 & 94.8 & 96.7 & 95.9 & 98.7 \\
        ViDoRe\_esg\_reports\_human\_labeled\_v2 & 63.4 & 6.8 & 45.6 & 50.7 & 50.4 & 60.6 & 68.0 & 61.8 & 64.0 \\
        ViDoRe\_biomedical\_lectures\_v2\_multilingual & 49.5 & 5.1 & 44.3 & 50.9 & 50.7 & 60.8 & 60.2 & 60.9 & 64.5 \\
        ViDoRe\_economics\_reports\_v2\_multilingual & 54.2 & 13.9 & 43.0 & 54.3 & 57.8 & 57.2 & 63.5 & 64.4 & 60.3 \\
        ViDoRe\_esg\_reports\_v2\_multilingual & 55.4 & 11.9 & 46.6 & 42.3 & 43.2 & 47.0 & 53.3 & 51.0 & 60.7 \\
        VisRAG\_ArxivQA & 87.4 & 52.6 & 76.9 & 74.0 & 80.5 & 76.6 & 81.3 & 84.5 & 88.0 \\
        VisRAG\_ChartQA & 81.9 & 70.2 & 84.4 & 82.7 & 85.0 & 86.4 & 86.5 & 86.8 & 89.0 \\
        VisRAG\_MP-DocVQA & 89.2 & 52.8 & 71.8 & 75.1 & 83.4 & 78.0 & 85.1 & 85.5 & 89.1 \\
        VisRAG\_SlideVQA & 94.5 & 72.8 & 91.5 & 87.6 & 91.5 & 90.3 & 95.1 & 95.4 & 96.0 \\
        VisRAG\_InfoVQA & 93.5 & 72.0 & 85.7 & 87.9 & 89.2 & 87.7 & 91.6 & 92.5 & 93.5 \\
        VisRAG\_PlotQA & 63.4 & 34.4 & 66.1 & 69.4 & 72.7 & 59.5 & 66.1 & 68.2 & 69.2 \\
        ViDoSeek-page & 23.2 & 22.3 & 21.9 & 22.5 & 21.3 & 23.0 & 21.9 & 22.4 & 23.1 \\
        ViDoSeek-doc & 83.9 & 77.8 & 80.2 & 73.8 & 75.3 & 80.8 & 83.5 & 83.3 & 83.4 \\
        MMLongBench-page & 16.2 & 11.8 & 11.9 & 13.3 & 12.3 & 14.9 & 15.1 & 15.0 & 15.8 \\
        MMLongBench-doc & 54.3 & 40.5 & 43.7 & 42.6 & 41.3 & 48.6 & 52.6 & 52.8 & 55.0 \\
        \bottomrule
    \end{tabular}
    \end{adjustbox}
\end{table*}

\section{Implementation Details}
\label{app:impl_more}

This section complements the main implementation setup in Sec.~\ref{subsubsec:imple_dets} by documenting practical details that are important for faithful reproduction beyond scalar hyperparameters.

\paragraph{Whitening and covariance regularization.}
We compute batch statistics from the concatenation of query and target embeddings, derive a single whitening transform, and apply it to both sets before regularizing second-order mismatch via a CORAL-style Frobenius penalty (Eq.~\ref{eq:coral_loss}).
To ensure numerical stability under high-dimensional embeddings, we implement the whitening operator with a robust routine that performs group-wise computation and injects a small jitter $\delta=10^{-4}$ into the covariance estimate when needed.

\paragraph{Debiased negative training.}
Our contrastive objective uses debiased contrastive learning to mitigate false-negative bias under negative selection.
In practice, we apply the debiasing term jointly with the curriculum-based masking strategy: early training avoids overly aggressive hard negatives, while later stages gradually increase negative pressure.

\paragraph{Hyperparameter selection protocol.}
For all hyperparameter sweeps reported in the paper, we tune on lightweight validation splits consisting of 1K samples drawn from the corresponding training sets, and report results on the MMEB-V2 test set(s) for consistency with prior work.

\section{Additional Hyperparameter Studies}
\label{app:param}

\begin{figure*}[t]
  \centering
  \includegraphics[width=0.9\textwidth]{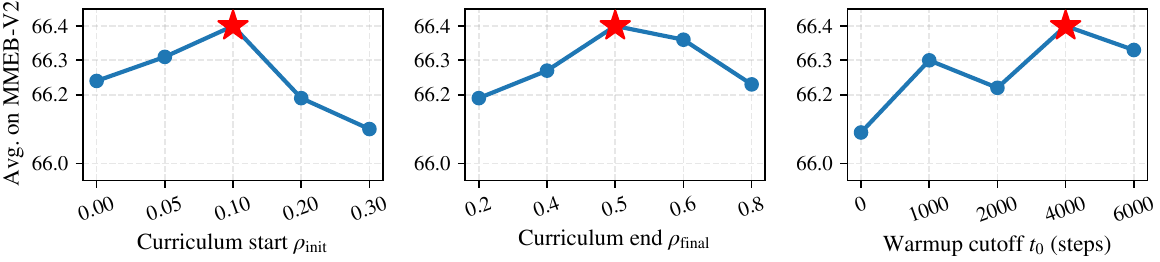}
  \caption{Sensitivity to negative curriculum settings on MMEB-V2 (7B backbone).}
  \label{fig:param_hns}
\end{figure*}

This appendix complements Sec.~\ref{subsec:param} with additional sweeps on the negative curriculum used for controllable negative selection (Sec.~\ref{subsec:method-hns}).
Unless otherwise specified, all sweeps are conducted on the 7B backbone with a fixed training budget.
We select hyperparameters on validation sets containing 1K samples drawn from the corresponding training sets, and report results on the MMEB-V2 test set for consistency with the main experiments (Fig.~\ref{fig:param_hns}).

\noindent\textbf{Curriculum start $\rho_{\mathrm{init}}$.}
We vary the initial quantile threshold $\rho_{\mathrm{init}}$, which controls how many negatives are kept at the beginning of training.
Very small $\rho_{\mathrm{init}}$ retains many easy negatives and yields weaker supervision, while overly large $\rho_{\mathrm{init}}$ can introduce hard (and potentially noisy) negatives too early.
As shown in Fig.~\ref{fig:param_hns}, performance peaks at a moderate start ($\rho_{\mathrm{init}}=0.10$) and drops on both sides, indicating that a balanced early hardness level is important.

\noindent\textbf{Curriculum end $\rho_{\mathrm{final}}$.}
We sweep the final threshold $\rho_{\mathrm{final}}$, which determines the hardness level near the end of training.
Increasing $\rho_{\mathrm{final}}$ improves performance up to $\rho_{\mathrm{final}}=0.50$, while more aggressive settings yield diminishing returns and eventually degrade performance.
We use $\rho_{\mathrm{final}}=0.50$ by default.

\noindent\textbf{Warmup cutoff $t_0$.}
Finally, we vary the warmup cutoff $t_0$ (in steps) before increasing hardness.
Starting the curriculum too early ($t_0=0$) underperforms, while moderate warmup improves stability and final accuracy.
We observe the best performance at $t_0=4000$ steps, and use $t_0=4000$ by default.

\end{document}